\documentclass{article}
\usepackage[table]{xcolor}
\usepackage{amsthm}
\usepackage{amsmath}
\usepackage{float}
\usepackage{lipsum}
\usepackage{amsfonts}
\restylefloat{figure}

\usepackage[usenames,dvipsnames]{pstricks}
\usepackage{epsfig}
\usepackage{pst-grad} 
\usepackage{pst-plot} 
\usepackage{graphicx} 
\usepackage{subfigure} 

\usepackage{natbib}

\usepackage{algorithm}
\usepackage{algorithmic}

\usepackage{hyperref}

\usepackage[accepted]{icml2013} 

\icmltitlerunning{Empirical Evaluation of Rectified Activations in Convolutional Network}

\begin{document} 

\twocolumn[
\icmltitle{Empirical Evaluation of Rectified Activations in Convolution Network}

\icmlauthor{Bing Xu}{antinucleon@gmail.com}
\icmladdress{University of Alberta}
\icmlauthor{Naiyan Wang}{winsty@gmail.com}
\icmladdress{Hong Kong University of Science and Technology}
\icmlauthor{Tianqi Chen}{tqchen@cs.washington.edu
}
\icmladdress{University of Washington}
\icmlauthor{Mu Li}{muli@cs.cmu.edu}
\icmladdress{Carnegie Mellon University}
\icmlkeywords{Deep Learning}

\vskip 0.3in
]

\begin{abstract} 

In this paper we investigate the performance of different types of rectified activation functions in convolutional neural network: standard rectified linear unit (ReLU), leaky rectified linear unit (Leaky ReLU), parametric rectified linear unit (PReLU) and a new randomized leaky rectified linear units (RReLU). We evaluate these activation function on standard image classification task. Our experiments suggest that incorporating a non-zero slope for negative part in rectified activation units could consistently improve the results. Thus our findings are negative on the common belief that sparsity is the key of good performance in ReLU. Moreover, on small scale dataset, using deterministic negative slope or learning it are both prone to overfitting. They are not as effective as using their randomized counterpart. By using RReLU, we achieved 75.68\% accuracy on CIFAR-100 test set without multiple test or ensemble. 

\end{abstract} 

\section{Introduction}
Convolutional neural network (CNN) has made great success in various computer vision tasks, such as image classification \citep{krizhevsky2012imagenet, szegedy2014going}, object detection\citep{rcnn} and tracking\citep{sodlt}. Despite its depth, one of the key characteristics of modern deep learning system is to use non-saturated activation function (e.g. ReLU)  to replace its saturated counterpart (e.g. sigmoid, tanh). The advantage of using non-saturated activation function lies in two aspects: The first is to solve the so called ``exploding/vanishing gradient". The second is to accelerate the convergence speed.

In all of these non-saturated activation functions, the most notable one is \emph{rectified linear unit} (ReLU)~\citep{nair2010rectified, sun2014deeply}. Briefly speaking, it is a piecewise linear function which prunes the negative part to zero, and retains the positive part. It has a desirable property that the activations are sparse after passing ReLU. It is commonly believed that the superior performance of ReLU comes from the sparsity~\citep{glorot2011deep, sun2014deeply}. In this paper, we want to ask two questions: \emph{First, is sparsity the most important factor for a good performance? Second, can we design better non-saturated activation functions that could beat ReLU?}


We consider a broader class of activation functions, namely the rectified unit family. In particular, we are interested in the leaky ReLU and its variants. In contrast to ReLU, in which the negative part is totally dropped, leaky ReLU assigns a noon-zero slope to it. The first variant is called \emph{parametric rectified linear unit} (PReLU) \citep{he2015delving}. In PReLU, the slopes of negative part are learned form data rather than predefined. The authors claimed that PReLU is the key factor of  surpassing human-level performance on ImageNet classification~\citep{ILSVRC15} task. The second variant is called \emph{randomized rectified linear unit} (RReLU). In RReLU, the slopes of negative parts are randomized in a given range in the training, and then fixed in the testing. In a recent Kaggle National Data Science Bowl (NDSB) competition\footnote{Kaggle National Data Science Bowl Competition: \url{https://www.kaggle.com/c/datasciencebowl}}, it is reported that RReLU could reduce overfitting due to its randomized nature.

In this paper, we empirically evaluate these four kinds of activation functions. Based on our experiment, we conclude on small dataset, Leaky ReLU and its variants are consistently better than ReLU in convolutional neural networks. RReLU is favorable due to its randomness in training which reduces the risk of overfitting. While in case of large dataset, more investigation should be done in future.

\section{Rectified Units}
\begin{figure}[!htb]
  \centering
    \includegraphics[width=0.5\textwidth]{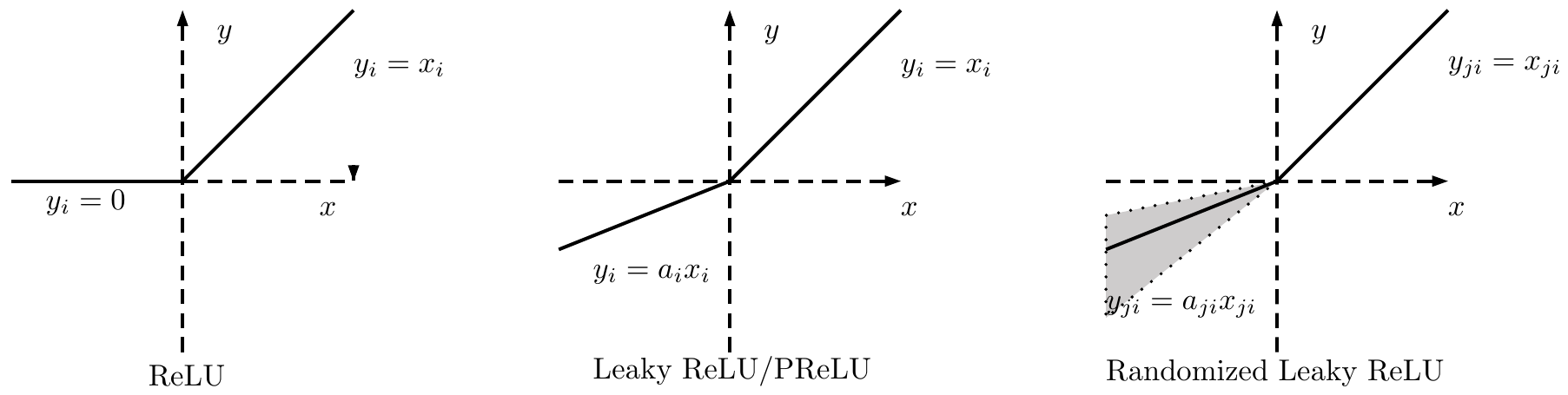}
      \caption{ReLU, Leaky ReLU, PReLU and  RReLU. For PReLU, $a_i$ is learned and for Leaky ReLU $a_i$ is fixed. For RReLU, $a_{ji}$ is a random variable keeps sampling in a given range, and remains fixed in testing.}
    \label{fig:act}
\end{figure}

In this section, we introduce the four kinds of rectified units: rectified linear (ReLU), leaky rectified linear (Leaky ReLU), parametric rectified linear (PReLU) and randomized rectified linear (RReLU). We illustrate them in Fig.\ref{fig:act} for comparisons. In the sequel, we use  $x_{ji}$ to denote the input of $i$th channel in $j$th example , and $y_{ji}$ to denote the corresponding output after passing the activation function. In the following subsections, we introduce each rectified unit formally.

\subsection{Rectified Linear Unit}
Rectified Linear is first used in Restricted Boltzmann Machines\citep{nair2010rectified}. Formally, rectified linear activation is defined as:

\begin{equation} \label{relu}
y_i=\left\{\begin{matrix}
x_i & \text{if } x_i \geq 0 \\
0 & \text{if } x_i < 0 .
\end{matrix}\right.
\end{equation}

\subsection{Leaky Rectified Linear Unit}
Leaky Rectified Linear activation is first introduced in acoustic model\citep{maas2013rectifier}. Mathematically, we have
\begin{equation} \label{lrelu}
y_i=\left\{\begin{matrix}
x_i & \text{if } x_i \geq 0 \\
\frac{x_i}{a_i} & \text{if } x_i < 0,
\end{matrix}\right.
\end{equation}
where $a_i$ is a fixed parameter in range $(1, +\infty)$. In original paper, the authors suggest to set $a_i$ to a large number like 100. In additional to this setting, we also experiment smaller $a_i = 5.5$ in our paper.
 
\subsection{Parametric Rectified Linear Unit}
Parametric rectified linear is proposed by \citep{he2015delving}. The authors reported its performance is much better than ReLU in large scale image classification task. It is the same as leaky ReLU (Eqn.\ref{lrelu}) with the exception that $a_i$ is learned in the training via back propagation.

\subsection{Randomized Leaky Rectified Linear Unit}
Randomized Leaky Rectified Linear is the randomized version of leaky ReLU. It is first proposed and used in Kaggle NDSB Competition. The highlight of RReLU is that in training process, $a_{ji}$ is a random number sampled from a uniform distribution $U(l, u)$. Formally, we have:
\begin{equation}
y_{ji}=\left\{\begin{matrix}
x_{ji} & \text{if } x_{ji} \geq 0 \\
a_{ji}x_{ji} & \text{if } x_{ji} < 0,
\end{matrix}\right.
\end{equation}
where
\begin{equation}
	a_{ji} \sim U(l, u) , l < u \text{ and } l, u \in [0, 1)
\end{equation}
In the test phase, we take average of all the $a_{ji}$ in training as in the method of dropout~\citep{srivastava2014dropout} , and thus set $a_{ji}$ to $\frac{l + u}{2}$ to get a deterministic result. Suggested by the NDSB competition winner, $a_{ji}$ is sampled from $U(3, 8)$. We use the same configuration in this paper.

In test time, we use:
\begin{align}
y_{ji} &= \frac{x_{ji}}{\frac{l + u}{2}}
\end{align}

\section{Experiment Settings}
We evaluate classification performance on same convolutional network structure with different activation functions. Due to the large parameter searching space, we use two state-of-art convolutional network structure and same hyper parameters for different activation setting. All models are trained by using CXXNET\footnote{CXXNET: \url{https://github.com/dmlc/cxxnet}}.

\subsection{CIFAR-10 and CIFAR-100}
The CIFAR-10 and CIFAR-100 dataset \citep{krizhevsky2009learning} are tiny nature image dataset. CIFAR-10 datasets contains 10 different classes images and CIFAR-100 datasets contains 100 different classes. Each image is an RGB image in size 32x32. There are 50,000 training images and 10,000 test images. We use raw images directly without any pre-processing and augmentation. The result is from on single view test without any ensemble. 

The network structure is shown in Table \ref{tab:nin}. It is taken from Network in Network(NIN)\citep{lin2013network}.

\rowcolors{2}{white}{gray!25}
\begin{table}[!htb]
\centering
    \begin{tabular}{ll}
    Input Size 		& NIN                  \\
    $32 \times 32$        & 5x5, 192             \\ 
    $32 \times 32$       		& 1x1, 160             \\ 
    $32 \times 32$        		& 1x1, 96              \\ 
    $32 \times 32$        		& 3x3 max pooling, /2  \\ 
    
    $16 \times 16$        		& dropout, 0.5         \\ 
    $16 \times 16$         		& 5x5, 192             \\ 
    $16 \times 16$         		& 1x1, 192             \\ 
    $16 \times 16$        		& 1x1, 192             \\ 
    $16 \times 16$         		& 3x3,avg pooling, /2  \\
    $8 \times 8$         		& dropout, 0.5         \\ 
    $8 \times 8$         		& 3x3, 192             \\ 
    $8 \times 8$        		& 1x1, 192             \\ 
    $8 \times 8$        		& 1x1, 10              \\
    $8 \times 8$        		& 8x8, avg pooling, /1 \\ 
    10 or 100       		& softmax             
    \end{tabular}
    \caption{CIFAR-10/CIFAR-100 network structure. Each layer is a convolutional layer if not otherwise specified. Activation function is followed by each convolutional layer.}
    \label{tab:nin}
\end{table}

In CIFAR-100 experiment, we also tested RReLU on Batch Norm Inception Network \citep{ioffe2015batch}. We use a subset of Inception Network which is started from inception-3a module. This network achieved 75.68\% test accuracy without any ensemble or multiple view test \footnote{CIFAR-100 Reproduce code: \url{https://github.com/dmlc/mxnet/blob/master/example/notebooks/cifar-100.ipynb}}.

\subsection{National Data Science Bowl Competition}
The task for National Data Science Bowl competition is to classify plankton animals from image with award of \$170k. There are 30,336 labeled gray scale images in 121 classes and there are 130,400 test data. Since the test set is private, we divide training set into two parts: 25,000 images for training and 5,336 images for validation. The competition uses multi-class log-loss to evaluate classification performance.

We refer the network and augmentation setting from team AuroraXie\footnote{Winning Doc of AuroraXie: \url{https://github.com/auroraxie/Kaggle-NDSB}}, one of competition winners. The network structure is shown in Table \ref{tab:ndsb}. We only use single view test in our experiment, which is different to original multi-view, multi-scale test.

\rowcolors{2}{white}{gray!25}
\begin{table}[!htb]
\centering
    \begin{tabular}{ll}
    Input Size & NDSB Net                  \\ 
    $70 \times 70$         & 3x3, 32                   \\ 
    $70 \times 70$         & 3x3, 32                   \\ 
    $70 \times 70$         & 3x3,  max pooling, /2     \\ 
    $35 \times 35$         & 3x3, 64                   \\ 
    $35 \times 35$         & 3x3, 64                   \\ 
    $35 \times 35$         & 3x3, 64                   \\ 
    $35 \times 35$         & 3x3, max pooling, /2      \\ 
    $17 \times 17$         & split: branch1 | branch 2 \\ 
    $17 \times 17$         & 3x3, 96 | 3x3, 96         \\ 
    $17 \times 17$         & 3x3, 96 | 3x3, 96         \\ 
    $17 \times 17$         & 3x3, 96 | 3x3, 96         \\ 
    $17 \times 17$         & 3x3, 96                   \\ 
    $17 \times 17$         & channel concat, 192       \\ 
    $17 \times 17$         & 3x3, max pooling, /2      \\ 
    $8 \times 8$          & 3x3, 256                  \\ 
    $8 \times 8$          & 3x3, 256                  \\ 
    $8 \times 8$          & 3x3, 256                  \\ 
    $8 \times 8$          & 3x3, 256                  \\ 
    $8 \times 8$          & 3x3, 256                  \\ 
    $8 \times 8$          & SPP~\citep{spp} \{1, 2, 4\}           \\ 
    $12544 \times 1$      & flatten                         \\ 
    $1024 \times 1$      & fc1                       \\ 
    $1024 \times 1$      & fc2                       \\ 
    121        & softmax                   \\ 
    \end{tabular}
    \caption{National Data Science Bowl Competition Network. All layers are convolutional layers if not otherwise specified. Activation function is followed by each convolutional layer.}
    \label{tab:ndsb}
\end{table}

\section{Result and Discussion}
\begin{figure*}[!htb]
\centering
\minipage{0.25\textwidth}
  \includegraphics[width=\linewidth]{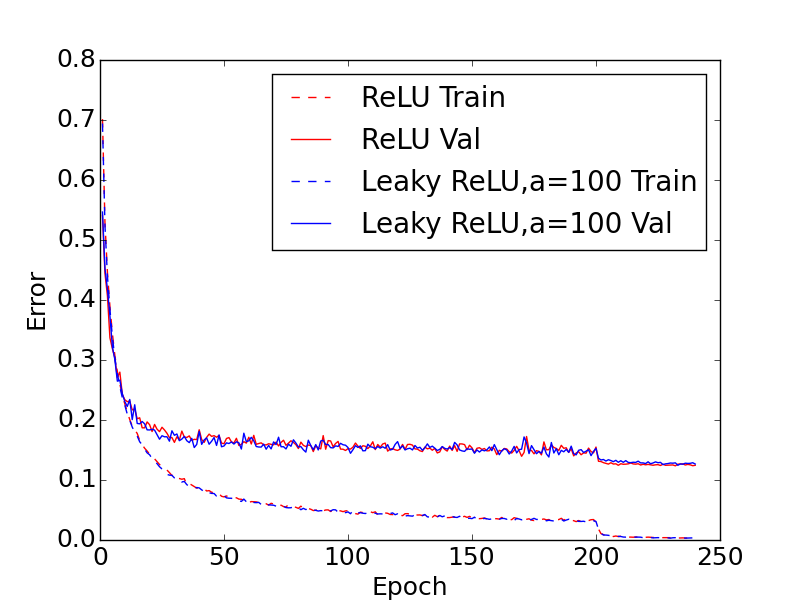}

\endminipage\hfill
\minipage{0.25\textwidth}
  \includegraphics[width=\linewidth]{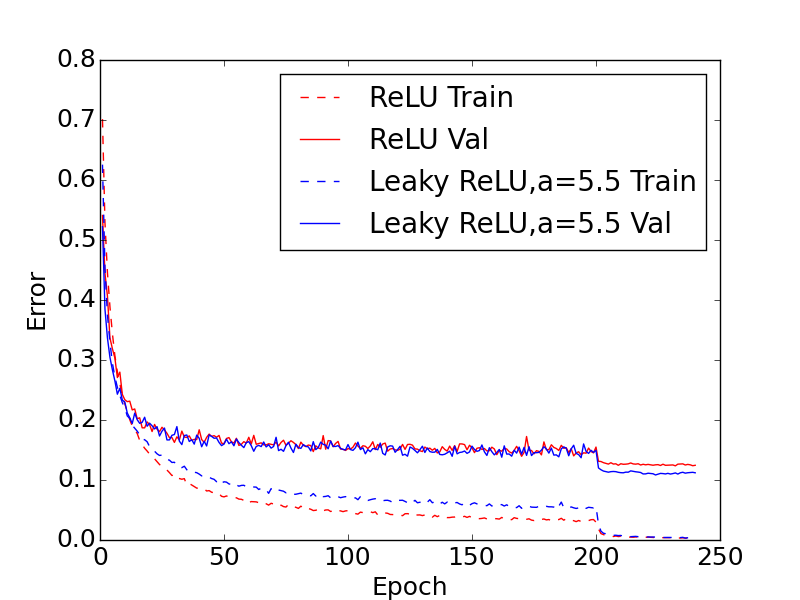}

\endminipage\hfill
\minipage{0.25\textwidth}%
  \includegraphics[width=\linewidth]{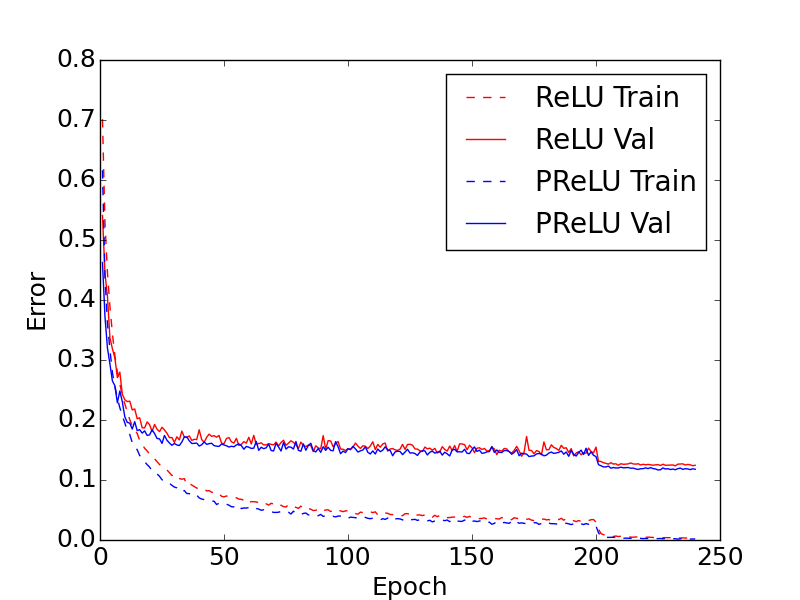}
\endminipage
\minipage{0.25\textwidth}
  \includegraphics[width=\linewidth]{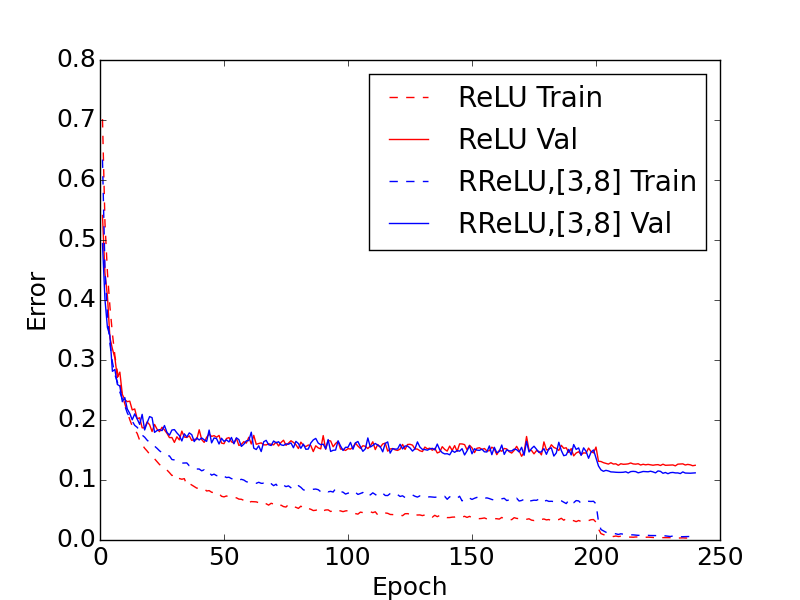}

\endminipage\hfill
\caption{Convergence curves for training and test sets of different activations on CIFAR-10 Network in Network.}
\label{fig:cifar}
\end{figure*}

\begin{figure*}[!htb]
\centering
\minipage{0.25\textwidth}
  \includegraphics[width=\linewidth]{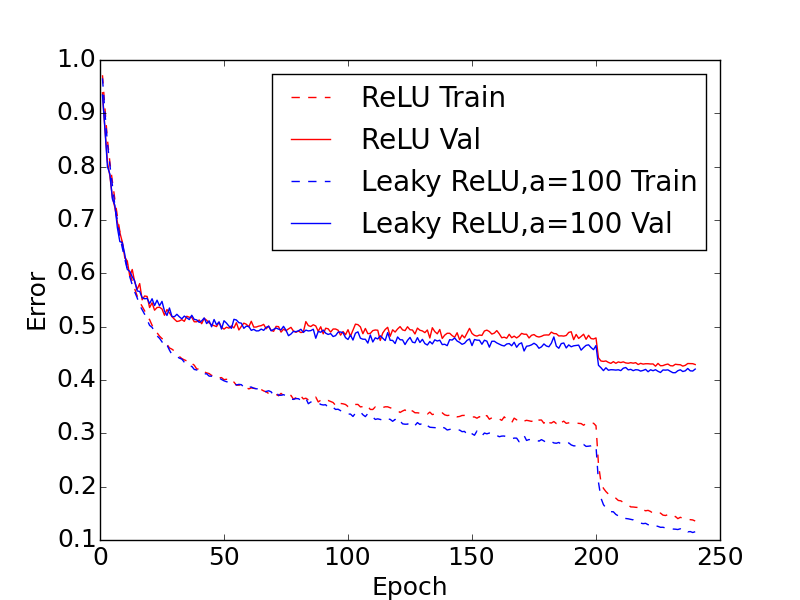}

\endminipage\hfill
\minipage{0.25\textwidth}
  \includegraphics[width=\linewidth]{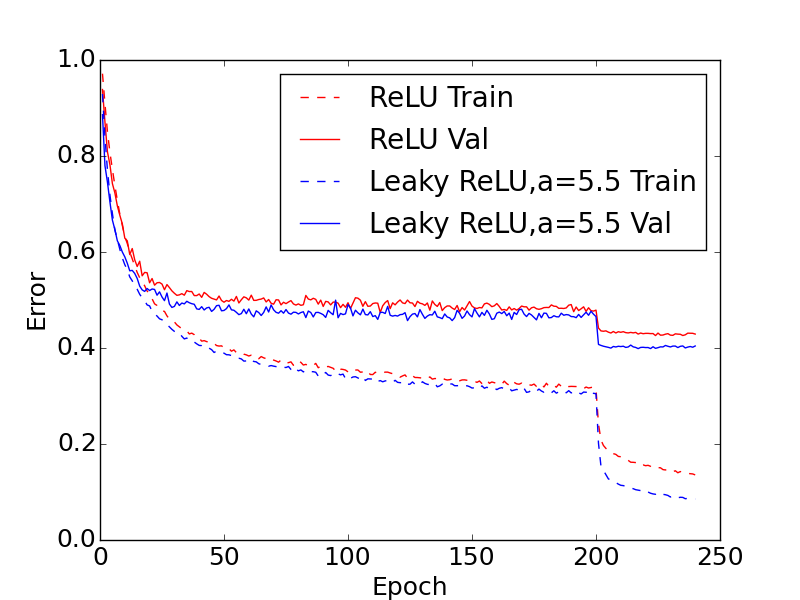}

\endminipage\hfill
\minipage{0.25\textwidth}%
  \includegraphics[width=\linewidth]{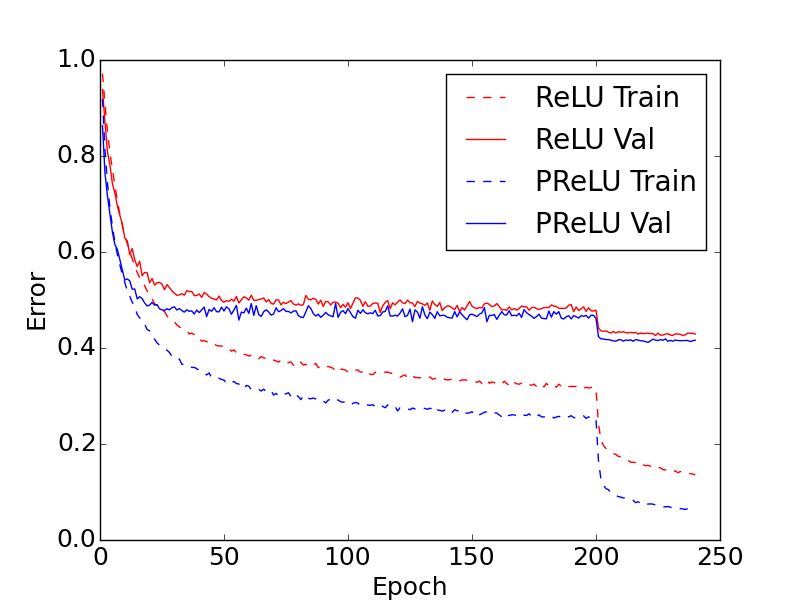}
\endminipage
\minipage{0.25\textwidth}
  \includegraphics[width=\linewidth]{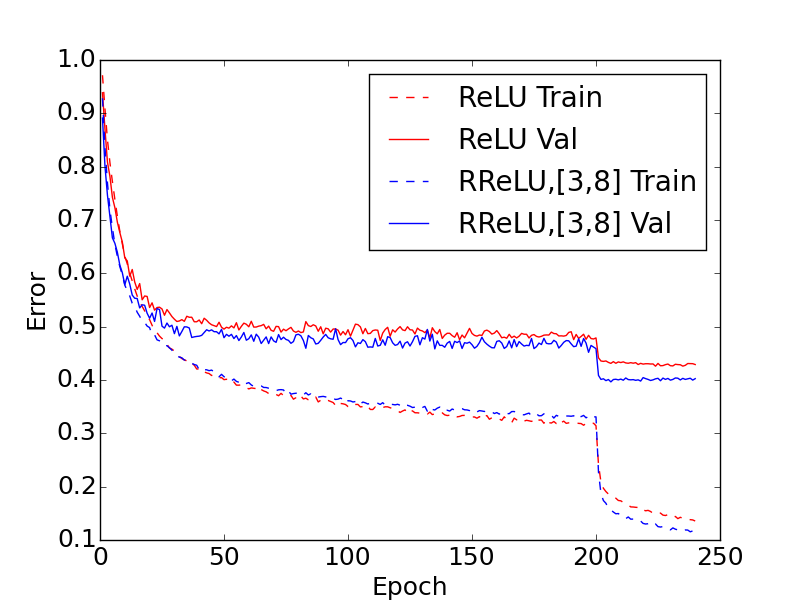}

\endminipage\hfill
\caption{Convergence curves for training and test sets of different activations on CIFAR-100 Network in Network.}
\label{fig:cifar100}
\end{figure*}

\begin{figure*}[!htb]
\centering
\minipage{0.25\textwidth}
  \includegraphics[width=\linewidth]{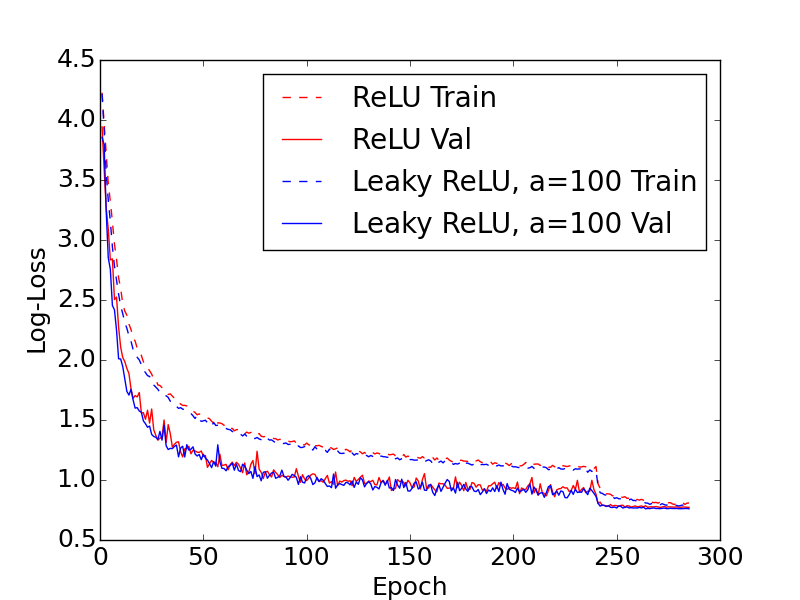}

\endminipage\hfill
\minipage{0.25\textwidth}
  \includegraphics[width=\linewidth]{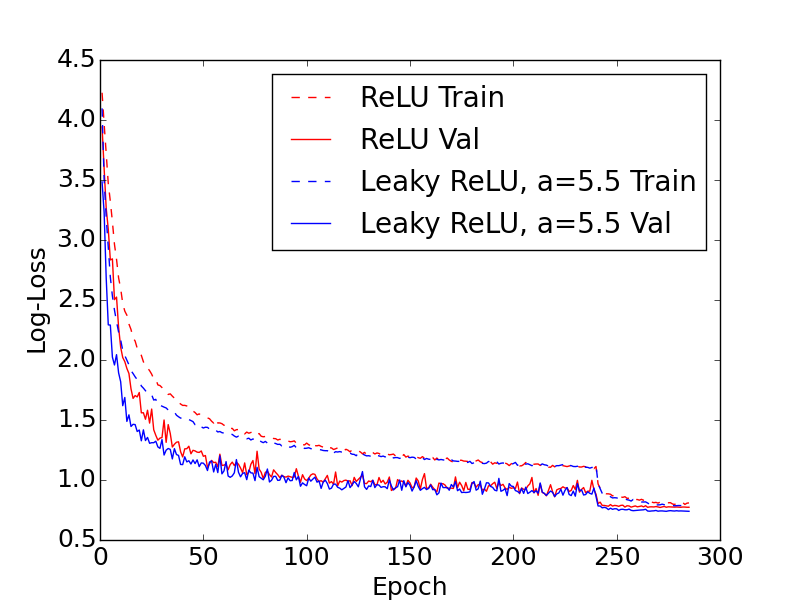}

\endminipage\hfill
\minipage{0.25\textwidth}%
  \includegraphics[width=\linewidth]{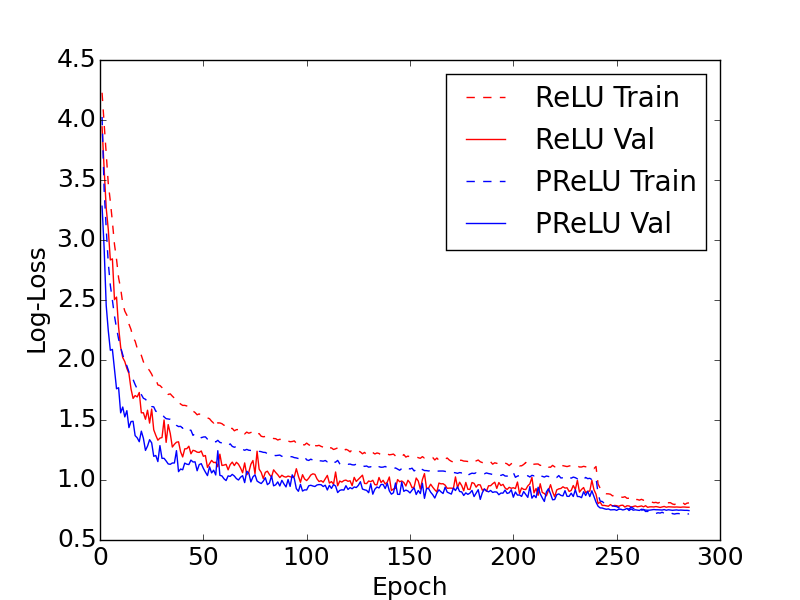}
\endminipage
\minipage{0.25\textwidth}
  \includegraphics[width=\linewidth]{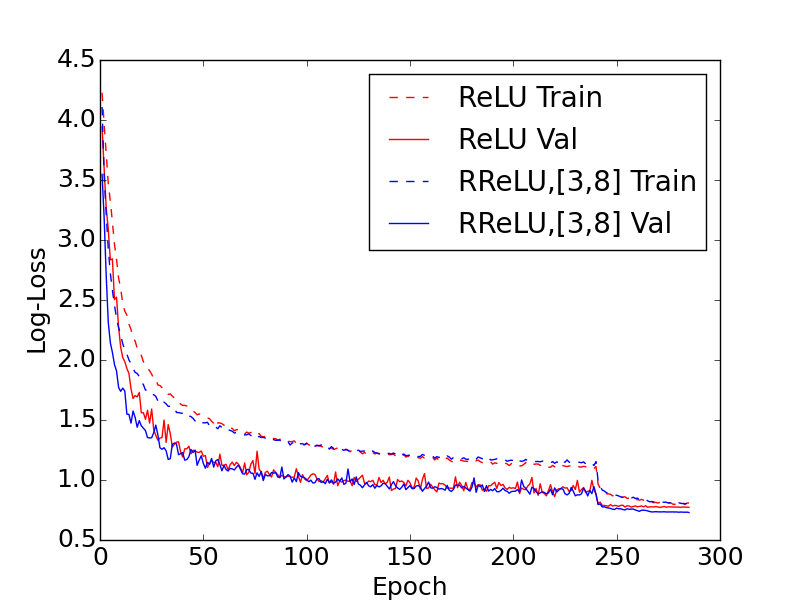}

\endminipage\hfill
\caption{Convergence curves for training and test sets of different activations on NDSB Net.}
\label{fig:ndsb}
\end{figure*}

Table \ref{tab:cifar} and \ref{tab:cifar100} show the results of CIFAR-10/CIFAR-100 dataset, respectively. Table \ref{tab:ndsb} shows the NDSB result. We use ReLU network as baseline, and compare the convergence curve with other three activations pairwisely in Fig. \ref{fig:cifar}, \ref{fig:cifar100} and \ref{fig:ndsb}, respectively. All these three leaky ReLU variants are better than baseline on test set. We have the following observations based on our experiment:
\begin{enumerate}
	\item  Not surprisingly, we find the performance of normal leaky ReLU ($a=100$) is similar to that of ReLU, but very leaky ReLU with larger $a = 5.5$ is much better. 
	\item On training set, the error of PReLU is always the lowest, and the error of Leaky ReLU and RReLU are higher than ReLU. It indicates that PReLU may suffer from severe overfitting issue in small scale dataset.
	\item The superiority of RReLU is more significant than that on CIFAR-10/CIFAR-100. We conjecture that it is because the in the NDSB dataset, the training set is smaller than that of CIFAR-10/CIFAR-100, but the network we use is even bigger. This validates the effectiveness of RReLU when combating with overfitting.
	\item For RReLU, we still need to investigate how the randomness influences the network training and testing process.
\end{enumerate}
\rowcolors{2}{white}{gray!25}
\begin{table}[!htb]
\centering
    \begin{small}
    \begin{tabular}{lll}
    Activation            & Training Error & Test Error \\ 
    ReLU                  & 0.00318        & 0.1245     \\ 
    Leaky ReLU, $a=100$            & 0.0031        & 0.1266     \\ 
    Leaky ReLU, $a=5.5$           & 0.00362        & \textbf{0.1120}     \\ 
    PReLU                 & 0.00178        & 0.1179     \\ 
    RReLU ($y_{ji} = x_{ji} /\frac{l+u}{2}$)  & 0.00550        & \textbf{0.1119}    
    \end{tabular}
    \end{small}
    \caption{Error rate of CIFAR-10 Network in Network with different activation function}
    \label{tab:cifar}
\end{table}

\rowcolors{2}{white}{gray!25}
\begin{table}[!htb]
\centering
    \begin{small}
    \begin{tabular}{lll}
    Activation            & Training Error & Test Error \\ 
    ReLU                  & 0.1356        & 0.429     \\ 
    Leaky ReLU, $a=100$            & 0.11552        & 0.4205     \\ 
    Leaky ReLU, $a=5.5$           & 0.08536        & \textbf{0.4042}     \\ 
    PReLU                 & 0.0633        & 0.4163     \\ 
    RReLU ($y_{ji} = x_{ji} /\frac{l+u}{2}$)  & 0.1141        & \textbf{0.4025}
    \end{tabular}
    \end{small}

    \caption{Error rate of CIFAR-100 Network in Network with different activation function}
    \label{tab:cifar100}
\end{table}

\rowcolors{2}{white}{gray!25}
\begin{table}[!htb]
\centering
    \begin{small}
    \begin{tabular}{lll}
    Activation & Train Log-Loss & Val Log-Loss \\ 
    ReLU       & 0.8092         & 0.7727       \\ 
    Leaky ReLU, $a=100$ & 0.7846              & 0.7601 \\ 
    Leaky ReLU, $a=5.5$ & 0.7831              & 0.7391 \\ 
    PReLU      & 0.7187              & 0.7454            \\ 
    RReLU ($y_{ji} = x_{ji} /\frac{l+u}{2}$)      & 0.8090         & \textbf{0.7292}  
    \end{tabular}
    \end{small}
    \caption{Multi-classes Log-Loss of NDSB Network with different activation function}
    \label{tab:ndsb}
\end{table}

\section{Conclusion}
In this paper, we analyzed four rectified activation functions using various network architectures on three datasets. Our findings strongly suggest that the most popular activation function ReLU is not the end of story: Three types of (modified) leaky ReLU all consistently outperform the original ReLU. However, the reasons of their superior performances still lack rigorous justification from theoretic aspect. Also, how the activations perform on large scale data is still need to be investigated. This is an open question worth pursuing in the future. 

\section*{Acknowledgement}
We would like to thank Jason Rolfe from D-Wave system for helpful discussion on test network for randomized leaky ReLU.
\bibliography{example_paper}

\begin{thebibliography}{15}
\providecommand{\natexlab}[1]{#1}
\providecommand{\url}[1]{\texttt{#1}}
\expandafter\ifx\csname urlstyle\endcsname\relax
  \providecommand{\doi}[1]{doi: #1}\else
  \providecommand{\doi}{doi: \begingroup \urlstyle{rm}\Url}\fi

\bibitem[Girshick et~al.(2014)Girshick, Donahue, Darrell, and Malik]{rcnn}
Girshick, Ross, Donahue, Jeff, Darrell, Trevor, and Malik, Jitendra.
\newblock Rich feature hierarchies for accurate object detection and semantic
  segmentation.
\newblock In \emph{CVPR}, pp.\  580--587, 2014.

\bibitem[Glorot et~al.(2011)Glorot, Bordes, and Bengio]{glorot2011deep}
Glorot, Xavier, Bordes, Antoine, and Bengio, Yoshua.
\newblock Deep sparse rectifier networks.
\newblock In \emph{Proceedings of the 14th International Conference on
  Artificial Intelligence and Statistics. JMLR W\&CP Volume}, volume~15, pp.\
  315--323, 2011.

\bibitem[He et~al.(2014)He, Zhang, Ren, and Sun]{spp}
He, Kaiming, Zhang, Xiangyu, Ren, Shaoqing, and Sun, Jian.
\newblock Spatial pyramid pooling in deep convolutional networks for visual
  recognition.
\newblock In \emph{ECCV}, pp.\  346--361, 2014.

\bibitem[He et~al.(2015)He, Zhang, Ren, and Sun]{he2015delving}
He, Kaiming, Zhang, Xiangyu, Ren, Shaoqing, and Sun, Jian.
\newblock Delving deep into rectifiers: Surpassing human-level performance on
  imagenet classification.
\newblock \emph{arXiv preprint arXiv:1502.01852}, 2015.

\bibitem[Ioffe \& Szegedy(2015)Ioffe and Szegedy]{ioffe2015batch}
Ioffe, Sergey and Szegedy, Christian.
\newblock Batch normalization: Accelerating deep network training by reducing
  internal covariate shift.
\newblock \emph{arXiv preprint arXiv:1502.03167}, 2015.

\bibitem[Krizhevsky \& Hinton(2009)Krizhevsky and
  Hinton]{krizhevsky2009learning}
Krizhevsky, Alex and Hinton, Geoffrey.
\newblock Learning multiple layers of features from tiny images.
\newblock \emph{Computer Science Department, University of Toronto, Tech. Rep},
  1\penalty0 (4):\penalty0 7, 2009.

\bibitem[Krizhevsky et~al.(2012)Krizhevsky, Sutskever, and
  Hinton]{krizhevsky2012imagenet}
Krizhevsky, Alex, Sutskever, Ilya, and Hinton, Geoffrey~E.
\newblock Imagenet classification with deep convolutional neural networks.
\newblock In \emph{NIPS}, pp.\  1097--1105, 2012.

\bibitem[Lin et~al.(2013)Lin, Chen, and Yan]{lin2013network}
Lin, Min, Chen, Qiang, and Yan, Shuicheng.
\newblock Network in network.
\newblock \emph{arXiv preprint arXiv:1312.4400}, 2013.

\bibitem[Maas et~al.(2013)Maas, Hannun, and Ng]{maas2013rectifier}
Maas, Andrew~L, Hannun, Awni~Y, and Ng, Andrew~Y.
\newblock Rectifier nonlinearities improve neural network acoustic models.
\newblock In \emph{ICML}, volume~30, 2013.

\bibitem[Nair \& Hinton(2010)Nair and Hinton]{nair2010rectified}
Nair, Vinod and Hinton, Geoffrey~E.
\newblock Rectified linear units improve restricted {Boltzmann} machines.
\newblock In \emph{ICML}, pp.\  807--814, 2010.

\bibitem[Russakovsky et~al.(2015)Russakovsky, Deng, Su, Krause, Satheesh, Ma,
  Huang, Karpathy, Khosla, Bernstein, Berg, and Fei-Fei]{ILSVRC15}
Russakovsky, Olga, Deng, Jia, Su, Hao, Krause, Jonathan, Satheesh, Sanjeev, Ma,
  Sean, Huang, Zhiheng, Karpathy, Andrej, Khosla, Aditya, Bernstein, Michael,
  Berg, Alexander~C., and Fei-Fei, Li.
\newblock {ImageNet Large Scale Visual Recognition Challenge}.
\newblock \emph{International Journal of Computer Vision (IJCV)}, 2015.
\newblock \doi{10.1007/s11263-015-0816-y}.

\bibitem[Srivastava et~al.(2014)Srivastava, Hinton, Krizhevsky, Sutskever, and
  Salakhutdinov]{srivastava2014dropout}
Srivastava, Nitish, Hinton, Geoffrey, Krizhevsky, Alex, Sutskever, Ilya, and
  Salakhutdinov, Ruslan.
\newblock Dropout: A simple way to prevent neural networks from overfitting.
\newblock \emph{The Journal of Machine Learning Research}, 15\penalty0
  (1):\penalty0 1929--1958, 2014.

\bibitem[Sun et~al.(2014)Sun, Wang, and Tang]{sun2014deeply}
Sun, Yi, Wang, Xiaogang, and Tang, Xiaoou.
\newblock Deeply learned face representations are sparse, selective, and
  robust.
\newblock \emph{arXiv preprint arXiv:1412.1265}, 2014.

\bibitem[Szegedy et~al.(2014)Szegedy, Liu, Jia, Sermanet, Reed, Anguelov,
  Erhan, Vanhoucke, and Rabinovich]{szegedy2014going}
Szegedy, Christian, Liu, Wei, Jia, Yangqing, Sermanet, Pierre, Reed, Scott,
  Anguelov, Dragomir, Erhan, Dumitru, Vanhoucke, Vincent, and Rabinovich,
  Andrew.
\newblock Going deeper with convolutions.
\newblock \emph{arXiv preprint arXiv:1409.4842}, 2014.

\bibitem[Wang et~al.(2015)Wang, Li, Gupta, and Yeung]{sodlt}
Wang, Naiyan, Li, Siyi, Gupta, Abhinav, and Yeung, Dit-Yan.
\newblock Transferring rich feature hierarchies for robust visual tracking.
\newblock \emph{arXiv preprint arXiv:1501.04587}, 2015.

\end{thebibliography}
\bibliographystyle{icml2013}

\end{document}